\pdfoutput=1

\documentclass[11pt]{article}

\usepackage[]{acl}

\usepackage{times}
\usepackage{latexsym}

\usepackage{graphicx}
\usepackage{booktabs}
\usepackage{subcaption}

\usepackage[T1]{fontenc}

\usepackage[utf8]{inputenc}

\usepackage{microtype}

\title{Measuring Entrainment in Spontaneous Code-switched Speech}

\author{Debasmita Bhattacharya \and Siying Ding \and Alayna Nguyen \and Julia Hirschberg \\
        Department of Computer Science \\ Columbia University \\ New York, NY, USA\\
        \texttt{\small{debasmita.b@cs.columbia.edu, sd3609@barnard.edu, amn2211@columbia.edu, julia@cs.columbia.edu}}}

\begin{document}
\maketitle
\begin{abstract}

It is well-known that speakers who entrain to one another have more successful conversations than those who do not. Previous research has shown that interlocutors entrain on linguistic features in both written and spoken \emph{monolingual} domains. More recent work on \emph{code-switched} communication has also shown preliminary evidence of entrainment on certain aspects of code-switching (CSW). However, such studies of entrainment in code-switched domains have been extremely few and restricted to human-machine textual interactions. Our work studies code-switched spontaneous speech between humans, finding that (1) patterns of written and spoken entrainment in monolingual settings largely generalize to code-switched settings, and (2) some patterns of entrainment on code-switching in dialogue agent-generated text generalize to spontaneous code-switched speech. Our findings give rise to important implications for the potentially "universal" nature of entrainment as a communication phenomenon, and potential applications in inclusive and interactive speech technology. 

\end{abstract}

\section{Introduction}

When people speak with one another, they often subconsciously adapt aspects of their communication style to that of their conversational partner. Interlocutors who do so perform what is known as \emph{entrainment} -- also called \emph{accommodation}, \emph{alignment}, or \emph{coordination}.  This has been shown to produce more successful conversations than conversations without entrainment \cite{reitter-moore-2007-predicting, nenkova-etal-2008-high}. 
In written and spoken domains of communication, people entrain  on multiple dimensions of language production: diction and syntax \cite{Danescu_Niculescu_Mizil_2011}, speaking rate, voice quality, and pause frequency \cite{giles_coupland_coupland_1991, levitan11_interspeech, chen2023}, jokes and laughter \cite{schmidt}, and facial expression and gesture \cite{maurer-tindall, burgoon, Chartrand1999TheCE}, among others. However, most of these findings come from studies on monolingual communication, which is much simpler to analyze than multilingual alternatives since there is no need to identify which language is being spoken for each word or phrase -- a challenging task applicable to most  multilingual corpora. 

\begin{figure}
\centering
\includegraphics[scale=0.44]{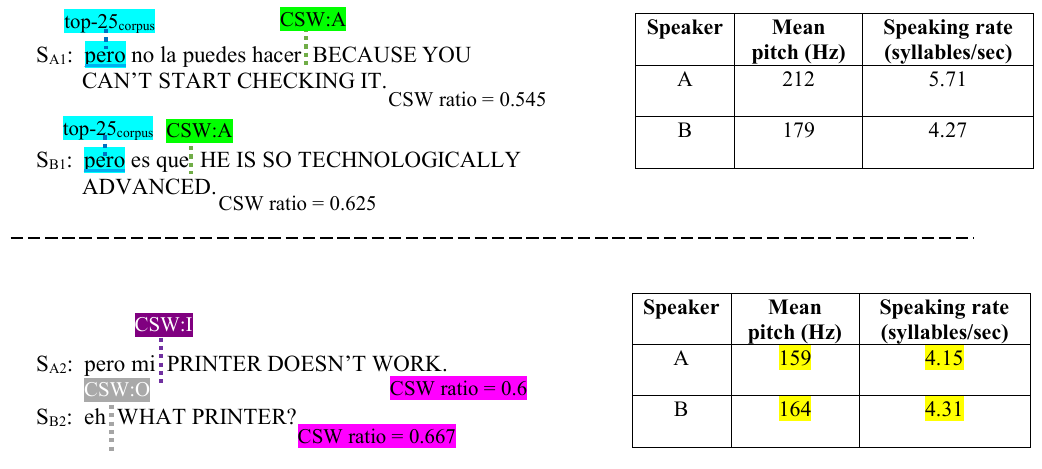}
    \caption{The novelty of our work comes from incorporating \textbf{acoustic-prosodic features} in the study of entrainment in \textbf{code-switched speech}. Here we highlight the value of identifying multiple dimensions and feature sets of entrainment in CSW. While entrainment is evident by inspection of the lexical and CSW strategy features of the $S_{A1}$, $S_{B1}$ interaction (top) (blue = entrainment on a frequent word; green = entrainment on alternational CSW), the acoustic-prosodic features are essential for correctly identifying entrainment in the $S_{A2}$, $S_{B2}$ interaction (bottom) (yellow = acoustic-prosodic entrainment; pink = entrainment on CSW amount; purple, gray = lack of entrainment on CSW strategy).}  
    \label{fig:1}
\end{figure}

This evidence of monolingual entrainment in prior research suggests that,  when multilingual speakers converse, they may also entrain to one another in aspects of multilingual language production such as code-switching (CSW), which occurs when a speaker alternates between one language and another in written or spoken communication \cite{poplack-1980}. There has been relatively little work on investigating entrainment in code-switched domains, particularly in speech production, leaving open the question of how  entrainment occurs in such multilingual settings. Answers to this question would be useful for understanding the conversations of the majority of speakers in the world who speak more than one language \cite{grosjean_li} and could inform future innovation in inclusive and interactive speech technology. 

We aim to fill this gap  by studying how entrainment manifests in naturalistic code-switched speech between humans. We first focus on Spanish-English CSW, which is common in Hispanic communities in the United States, by examining the lexical, acoustic-prosodic, and multilingual characteristics of the Bangor Miami code-switched conversations \cite{miami} (Figure~\ref{fig:1}). Our main contributions from doing so are (1) adapting metrics of entrainment previously used in monolingual settings to multilingual ones. We study lexical and acoustic-prosodic entrainment in a code-switched setting, and find evidence that entrainment occurs in multilingual domains just as it does in monolingual ones. (2) We identify CSW patterns of entrainment in spontaneous human-human speech 
to inform the design of interactive speech technology capable of both understanding and appropriately responding to code-switched speech in a human-like way. (3) We also create an updated version of the original Bangor Miami corpus, where the audio data is cleaned of background noise and the textual data includes new annotations. We make this updated corpus publicly available to the research community.\footnote{\url{https://drive.google.com/drive/folders/1Ty2WzGwSiolDeN7J9yK5r6XCoMrUowQa?usp=sharing}} 

\section{Related work and Research questions}
\textbf{Entrainment in monolingual domains. }  Much  work has been done on entrainment in text and speech in monolingual domains. \citet{levitan11_interspeech} proposed an early framework for measuring entrainment in speech, distinguishing entrainment at the turn-level from entrainment at the conversation-level and identifying multiple ways of entraining, including \emph{ proximity}, \emph{convergence}, and \emph{synchrony}. Separate from this  framework for measuring  acoustic-prosodic entrainment are count-based and probabilistic measures applied often in studies of lexical entrainment. \citet{nenkova-etal-2008-high} introduced several relative, symmetric metrics for computing lexical entrainment over word classes, which were applied and extended in \citet{weise-etal-2021-talk}. \citet{gravano_tobi} studied prosodic entrainment over ToBI elements, using a perplexity-based method for calculating asymmetric coordination on the textual form of prosodic features. Greater evidence of entrainment has been correlated with greater task success and improved social outcomes \cite{reitter-moore-2007-predicting, levitan-etal-2012-acoustic}. Analysis of entrainment patterns in relation to speakers' demographic characteristics has revealed interactions with gender dynamics \cite{BILOUS1988183, levitan-etal-2012-acoustic, cabarrao} and power differentials \cite{Danescu_Niculescu_Mizil_2011}. Although some work has studied entrainment in languages other than English, few have examined entrainment in multilingual contexts where CSW occurs, giving rise to our first research question: 
\textbf{RQ1:} Do previously established patterns of entrainment in monolingual settings generalize to code-switched settings -- is entrainment a universal phenomenon of spoken communication? Do language behaviors  found in monolingual communication also occur in code-switched communication?

\textbf{Entrainment for CSW.}  Recent work on entrainment has begun to explore code-switched domains, but most studies have been restricted to the textual modality. \citet{KOOTSTRA2010210} found that Dutch-English speakers entrain on CSW more when their interlocutor has just code-switched, and that they use shared word order in order to code-switch then, focusing entirely on lexical entrainment. \citet{soto18_interspeech} examined only convergence of the amount of CSW that occurs between two Spanish-English speakers over the course of a conversation, analyzing speech transcripts. Similarly, \citet{ahn-etal-2020-code} and \citet{parekh-etal-2020-understanding} studied entrainment of CSW \emph{strategies} in human-computer communication in Spanish-English and Hindi-English respectively, distinguishing between relatively simple \emph{insertional} (inserting a single code-switched word or short phrase into an otherwise monolingual utterance, e.g.~\emph{Todos estamos con un calor y WORKING.}) and more complex \emph{alternational} (CSW occurring at grammatical clause boundaries, e.g.~\emph{No tienes que pagar mucho porque YOU DON'T HAVE THE CHECK.}) code-switches. Given the nature of human-machine dialogue systems, this work was also restricted to analysis of written interactions. The findings of these studies beg the question: how does entrainment manifest in naturalistic code-switched speech between humans, leading to our second research question: \textbf{RQ2: } Do patterns of entrainment in code-switched text (some of which is produced by virtual dialogue agents) generalize to code-switched spontaneous speech?  Do code-switched language behaviors depend on the modality of language production and/or the nature of the interlocutor, and are there similarly generalizable characteristics of CSW that are independent of these conversational factors? 



\section{Corpus}

We examine the \href{ http://bangortalk.org.uk}{Bangor Miami} corpus of informal conversations \cite{miami}.\footnote{This corpus is made available under the \href{http://gnu.org/copyleft/gpl.html}{GNU General Public
License} version 3 or later.} This corpus is composed of a mix of monolingual and code-switched English and Spanish spontaneous speech, along with transcripts of these, as produced by 84 unique bilingual Spanish-English speakers living in Miami, Florida. The corpus contains 35 hours of recorded conversation and 242,475 words of transcribed text. The transcribed text comes manually annotated with word-level language identification labels for each utterance in each conversation. The complete corpus consists of 56 conversations, with the number of speakers per conversation ranging between one and four. Of these, 39 are dyadic conversations (i.e. conversations that involve exactly two speakers) that contain code-switched utterances.  We restrict our investigation to the 20 hours of these conversations only. 

\section{Method}


\textbf{Preprocessing: data annotation.} We begin by annotating the Bangor Miami corpus to identify the different CSW strategies used by speakers in the filtered corpus of 39 dyadic conversations. We accomplish this by inspection of the conversation transcripts. We first automatically label each utterance of each conversation's transcript for whether it is code-switched or monolingual, based on its word-level language tags. Two annotators fluent in Spanish and English then perform additional manual annotations on the code-switched utterances to distinguish between insertional (\emph{I}), alternational (\emph{A}), and "other" (\emph{O}) forms of CSW.\footnote{When the annotators disagree on a label, they discuss their reasoning with each other until the disagreement is resolved.} 
The annotators define "other" CSW as the strategy used in any utterance where a code-switched filler word such as \emph{okay} or \emph{pues} appears at the outset or end of a sentence whose remainder is monolingual in the opposite language. We find that 95\% of utterances in the filtered corpus are monolingual. Among the 5\% of utterances in the corpus that are code-switched, 72\% of these use insertional CSW, 13\% use alternational CSW, and 18\% use "other" CSW (some code-switched utterances use more than one CSW strategy).

\textbf{Preprocessing: denoising audio data.} Next, we handle the babble noise (i.e. the kind of background noise often heard in restaurants, bars, or airports)
present in most of the corpus' audio files, to ensure reliable downstream feature extraction and analysis. To achieve this, we create a denoising pipeline. This is a two-step process that consists of first applying to each audio file a Conformer-based Metric Generative Adversarial Network \cite{Cao_2022}, which enhances the magnitude and spectrogram features of a speech signal, followed by using \href{https://manual.audacityteam.org/man/noise_reduction.html}{Audacity’s} publicly-available noise reduction tools. This approach effectively reduces background noise while preserving the integrity of the speech signal,\footnote{Note that we conducted spot checks during data preprocessing where the first three authors listened to a subset of individual cleaned audio files, visually compared noisy and denoised waveforms for consistency, and ensured that extracted acoustic-prosodic features were consistent with the waveforms before and after denoising.} as evidenced by our calculated speech-to-noise (SNR) ratio in each denoised audio file exceeding 30 dB, the threshold for clean speech signals \cite{SNR}. The mean, mode, and median SNR in the corpus post-denoising are 54.3dB, 74.7dB, and 56.3dB respectively. These are significant improvements over the pre-denoising mean, mode, and median SNR of 29.3dB, 35.6dB, and 29.3dB respectively. We share this newly cleaned version of the corpus with the spoken language processing community.

\textbf{Measuring entrainment across feature sets.} Following the preprocessing of the corpus, we calculate several measures of entrainment across lexical, acoustic-prosodic, and CSW feature sets. 
The \textbf{lexical feature set} includes the following word classes, as in \citet{nenkova-etal-2008-high}: \textbf{most frequent words within the corpus} (top-100 and top-25), \textbf{most frequent words within each conversation} (top-25), \textbf{affirmative cue words} (e.g. \emph{alright}, \emph{gotcha}, \emph{okay}, \emph{uh-huh}, \emph{yeah}, and their respective Spanish equivalents), and \textbf{filled pauses} (e.g. \emph{uh}, \emph{um}, \emph{mm}, and their respective Spanish equivalents). See Appendix~\ref{sec:appendix} for the full list of affirmative cue words and filled pauses considered. Note that both Spanish and English words are members of each of the above word classes. Our experiments on these lexical features of the corpus use Equation~\ref{eq1} to calculate entrainment in each of the word classes $W$ under investigation.

\texttt{\small{\begin{equation}
entr(S_A, S_B)=-\sum_{w \in W}|\frac{count_{S_A}(w)}{ALL_{S_A}}-\frac{count_{S_B}(w)}{ALL_{S_B}}|
\label{eq1}
\end{equation}}}

This equation was first proposed in \citet{nenkova-etal-2008-high}. It defines entrainment between two speakers $S_A$ and $S_B$ on a particular word class $W$ as the negated absolute value of the difference between the fraction of times a particular word $w \in W$ is used by $S_A$ and $S_B$, summed over all the words in $W$.

We also compute \textbf{lexical entrainment on overall language use} in the corpus, adopting the method used by \citet{gravano_tobi}. We use the speech transcripts of each speaker as training data for estimating a Kneser-Essen-Ney smoothed trigram model, using the KenLM toolkit for this task, and resulting in 78 such trigram models --- two for each conversation in the corpus. For each conversation, we evaluate the trained model for $S_A$ by computing its perplexity on the speech transcripts of $S_B$. The negated perplexity value produces an entrainment score reflecting how much $S_A$ entrains to $S_B$. We do the same perplexity calculations in the opposite direction to obtain an entrainment score for how much $S_B$ entrains to $S_A$. The lower the perplexity, the greater the entrainment.  

The members of the \textbf{acoustic-prosodic feature set} for each utterance are: \textbf{minimum, mean, maximum, and standard deviation in pitch}; \textbf{minimum, mean, maximum, and standard deviation in intensity}; \textbf{jitter}; \textbf{shimmer}; \textbf{harmonics-to-noise ratio} (HNR); and \textbf{speaking rate}, measured in syllables per second. We extract these acoustic-prosodic features automatically using the Parselmouth Python library \cite{parselmouth} for the Praat software \cite{praat}, with all parameters set to their default values.  We define the \textbf{set of CSW features} such that, for each utterance, we examine the \textbf{binary presence of CSW} (coded as 0 for monolingual utterances or 1 for code-switched ones), the \textbf{amount of CSW} (normalized by the word length of the utterance; this produces a CSW \emph{ratio}), and the \textbf{strategy of CSW} used (\emph{I}, \emph{A}, or \emph{O} for code-switched utterances; -1 for monolingual utterances). For our experiments on the acoustic-prosodic and CSW feature sets, we calculate entrainment in terms of \textbf{proximity} (the absolute similarity of a feature over an entire turn/conversation), \textbf{convergence} (the degree to which a feature becomes more similar over the course of a conversation), and \textbf{synchrony} (turn-by-turn \emph{relative} coordination between interlocutors, as when speakers have different pitch ranges but raise and lower their pitch similarly to their partner) at the turn- and conversation-level, following the methods used in \citet{levitan11_interspeech}.

For \textbf{turn-level proximity}, we perform calculations in the following way. For each target speaker turn, we compute a partner difference (Equation~\ref{eq2}) and other difference (Equation~\ref{eq3}), such that $turn_p$ is adjacent to the target turn and uttered by the target turn speaker’s conversational partner, and $turn_i$ is uttered by the target turn speaker’s conversational partner but is not adjacent to the target turn, for ten random turns. We compare partner differences and other differences with a paired \emph{t}-test. We infer proximity when partner differences from the prior speaker turn are smaller than  differences from other speaker turns.

\texttt{\small{\begin{equation}
    difference_{partner} = |turn_t - turn_p|
\label{eq2}
\end{equation}}}
\texttt{\small{\begin{equation}
    difference_{other} = \frac{\sum_{i=1}^{10}|turn_t - turn_i|}{10}
\label{eq3}
\end{equation}}}

For \textbf{turn-level convergence}, we compute the Pearson's correlation coefficient between the absolute value of the difference between adjacent turns and the turn number, which effectively measures time elapsed in the conversation. We infer strong, moderate, or weak convergence from Pearson's coefficients $r \geq 0.7$, $0.5 \leq r<0.7$, and $0<r<0.5$, respectively. We infer strong, moderate, and weak divergence from Pearson's coefficients $r \leq -0.7$, $-0.7<r \leq -0.5$, and $-0.5<r<0$, respectively.

For \textbf{turn-level synchrony}, we compute the Pearson's correlation coefficient between adjacent turns from different speakers, testing for significance with a two-sided \emph{t}-test. We infer strong, moderate, and weak synchrony from Pearson's coefficients $r \geq 0.7$, $0.5 \leq r<0.7$, and $0<r<0.5$, respectively. We infer strong, moderate, and weak asynchrony from Pearson's coefficients $r \leq -0.7$, $-0.7<r \leq -0.5$, and $-0.5<r<0$, respectively.

For \textbf{conversation-level proximity}, we use paired \emph{t}-tests on two sets of differences. For each speaker, we calculate a partner difference (the difference between the speaker's value for a particular feature and that of their partner) and an other difference (the mean of the differences between the speaker’s value and the values of each speaker in the corpus who was not their interlocutor). We infer proximity when partner differences are smaller than other differences.

For \textbf{conversation-level convergence}, we split each conversation into two halves. In each half, we calculate each speaker's mean value for a particular feature. We then compare  differences in speaker mean values in the first half to their differences in mean values in the second half using a paired \emph{t}-test. We infer convergence when the differences in the second half are significantly smaller.

We perform \emph{z}-score normalization by speaker on all acoustic-prosodic features prior to any entrainment calculation to account for each individual speaker's natural pitch range and voice qualities. 

\section{Experiments}

\subsection{Interlocutors entrain on lexical features of CSW conversations}

We begin our investigation of \textbf{lexical entrainment} on the conversations of the Bangor Miami corpus by using Equation~\ref{eq1} to compute a single entrainment score for each conversation, aggregated over each \textbf{word class}. We test for statistical significance using paired \emph{t}-tests for each word class and compare entrainment scores between conversational partners to mean entrainment scores between non-partners. For the top-100 words within the corpus and affirmative cues, all 39 conversations in the corpus show significant evidence of lexical entrainment ($t=19.2; p=4.50e-31$ and $t=8.26; p=3.21e-12$, respectively). For the top-25 words within the corpus, most frequent words within each conversation, and filled pauses, 38 conversations in the corpus show significant evidence of lexical entrainment ($t=18.6, p=3.14e-30$; $t=13.4, p=7.20e-22$; $t=8.84, p=2.42e-13$, respectively). 



Having considered several pre-defined word classes, we next investigate entrainment on \textbf{overall language use}. 
When including out-of-vocabulary (OOV) words in our calculations, we find that all but one conversation in the corpus 
shows evidence of each speaker entraining to their partner's overall language use as quantified by perplexity.  We test the significance of this result using a paired \emph{t}-test to compare perplexity scores between conversational partners to the mean of perplexity scores between non-partners for each speaker in the corpus, yielding $t=-11.9$ and $p=3.58e-19$. When excluding OOV words in our calculations, we find that 68 of out 78 possible $(conversation, partner)$ combinations show evidence of within-conversation entrainment on overall language use. In other words, in 29 out of 39 conversations both speakers are entraining to their conversational partner, while in the remaining 10 conversations only one speaker is entraining to their interlocutor. As above, we test the significance of this result using a paired \emph{t}-test, yielding $t=-8.73$ and $p=4.06e-13$. 

\subsection{Interlocutors entrain on most acoustic-prosodic features of CSW conversations}

\begin{figure}
\includegraphics[scale=0.38]{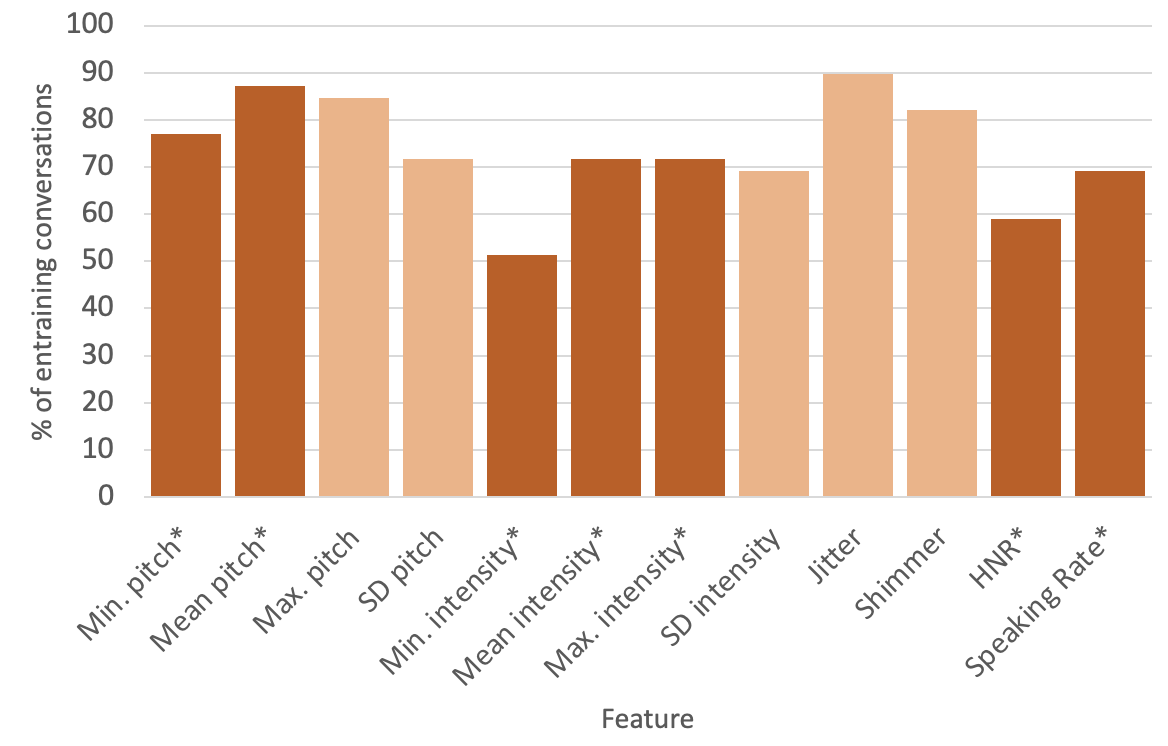}
    \caption{Proximity at the turn-level. Significant acoustic-prosodic features are indicated by * and dark orange bars. See Table~\ref{tab:proximity-turn} in Appendix~\ref{sec:appendix} for \emph{t} and \emph{p}-values corresponding to each feature.}  
    \label{proximity-turn}
\end{figure}

\begin{figure}
\includegraphics[scale=0.38]{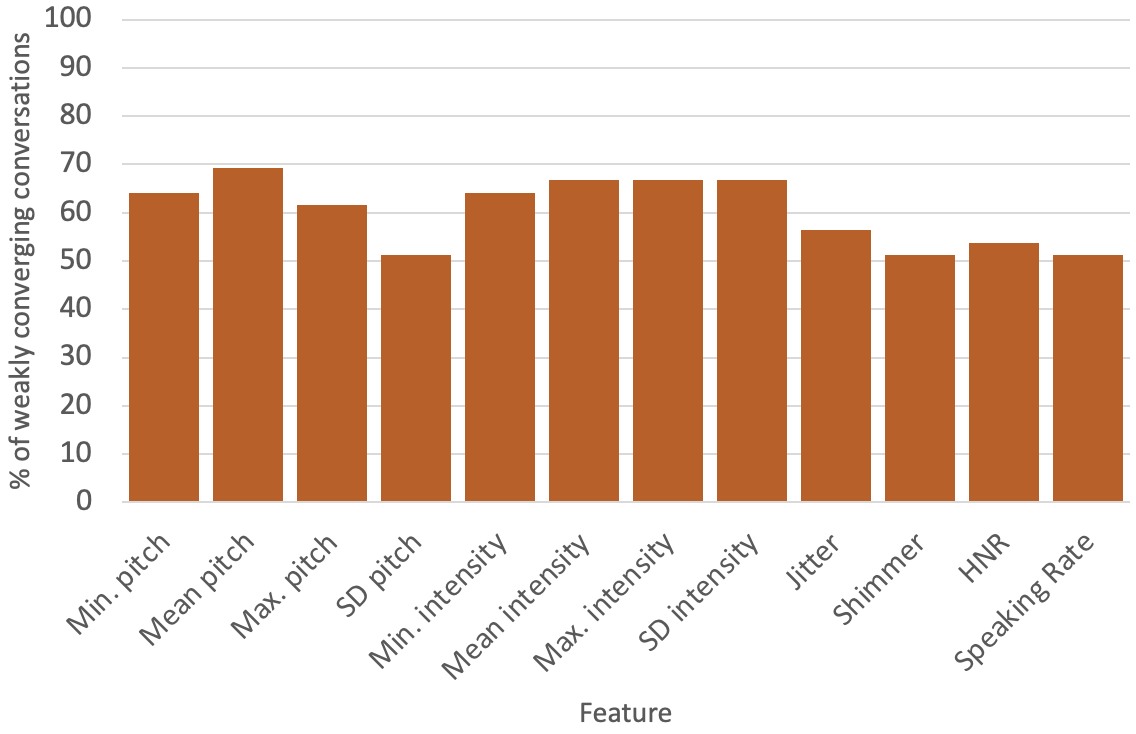}
    \caption{Convergence at the turn-level. All acoustic-prosodic features are significant. The percentage of weakly diverging conversations for each acoustic-prosodic feature is the difference between the percentage of weakly converging conversations and 100.}  
    \label{convergence-turn}
\end{figure}

Following our investigation of lexical features, we perform experiments to measure \textbf{acoustic-prosodic entrainment} on the conversations of the Bangor Miami corpus. 
In terms of \textbf{proximity at the turn-level}, we find that the majority of conversations show statistically significant evidence of entrainment in terms of proximity on the majority of acoustic-prosodic features (Figure~\ref{proximity-turn}). Similarly, in terms of \textbf{convergence at the turn-level}, the majority of conversations show statistically significant evidence of entrainment in the same direction on all of the acoustic-prosodic features under investigation (Figure~\ref{convergence-turn}). However, we do not find significant evidence of turn-level synchrony nor conversation-level proximity or convergence of the acoustic-prosodic features.

Since we mainly find evidence of \emph{turn-level} entrainment on the acoustic-prosodic features, we additionally compare entrainment between turns at the start, middle, and end of conversations. Doing so, we find greater entrainment in terms of turn-level proximity in the initial third of conversations (i.e. greater number of conversations satisfying entrainment conditions) compared to the final third of conversations; these results are significant for minimum, mean, and maximum pitch, and maximum intensity. We also find greater entrainment in terms of turn-level convergence in both the initial and middle third of conversations (i.e. greater Pearson correlation coefficients) compared to the final third of conversations, across acoustic-prosodic features, all of which are significant. These more granular turn-level results align with our expectations and previous conversation-level convergence results, as greater entrainment in earlier parts of a conversation compared to later parts indicates an absence of conversation-level convergence by definition.

\subsection{Interlocutors entrain on most CSW features of CSW conversations}

Our final set of quantitative experiments measure entrainment in terms of a number of \textbf{CSW characteristics} of the conversations of the Bangor Miami corpus. We start with \textbf{entrainment on the presence of CSW}. In terms of \textbf{proximity at the turn-level}, we find that 29 of the 39 conversations in the corpus show evidence of entrainment on the presence of CSW. A paired \emph{t}-test shows that this is statistically significant ($t=-2.420$; $p=0.0155$). We similarly find that 21 conversations are weakly \textbf{synchronous at the turn-level} (two-sided \emph{t}-test $p =0.0001$), and 32 conversations show \textbf{proximity at the conversation-level} ($t=-3.402$; $p=0.001$). However, we find limited evidence of convergence at the turn- or conversation-level, likely because speakers entrain in proximity quite rapidly.

Next, we turn to \textbf{entrainment on the amount of CSW} used in the Bangor Miami conversations. We find \textbf{proximity at the turn-level}, where 33 conversations show statistically significant evidence of entrainment on the amount of CSW performed ($t=-2.153$; $p=0.0314$), and \textbf{proximity at the conversation-level}, where 34 conversations show significant evidence of entrainment ($t=-2.470$; $p=0.0157$). We also find that 1 conversation is moderately synchronous and 21 others are weakly synchronous, however this result is not statistically significant ($p=0.590$). As above, we find no significant evidence of turn- or conversation-level convergence on amount of CSW, as these conversational partners entrain in proximity quite rapidly. 

Finally, we consider \textbf{entrainment on CSW strategies}. In this set of experiments, we perform separate calculations for each possible strategy (insertional, alternational, or "other" CSW). We find \textbf{proximity at the conversation-level} on alternational CSW only, with 29 conversations showing statistically significant evidence of entrainment on this particular strategy ($t= -3.378$; $p=0.001$). We also find significant \textbf{synchrony at the turn-level} on insertional CSW, with 22 weakly synchronous conversations ($p=1.528e-05$). We do not find significant evidence of turn-level proximity or convergence nor conversation-level convergence for any of the CSW strategies. As a whole, our set of results on strategies of CSW is largely consistent with those on the other characteristics of code-switched conversations.

As with the acoustic-prosodic features, we compare entrainment between turns at the start, middle, and end of conversations. We find greater entrainment in terms of turn-level proximity and synchrony in the middle third of conversations compared to the final third of conversations for CSW presence and CSW amount. We also find greater entrainment in terms of turn-level synchrony in both the initial and middle third of conversations compared to the final third of conversations across all CSW strategies, all of which at least approach significance. As before, these granular turn-level results are reasonable as they align with a lack of conversation-level convergence on CSW features. 

\subsection{Same-gender interlocutors entrain more than opposite-gender interlocutors}
We finally perform a qualitative analysis of our results in relation to the genders of the speakers in conversations that showed significant evidence of entrainment. We particularly distinguish between same-gender (FF or MM) conversations and mixed-gender (FM) conversations. Our comparisons are weighted to account for there being 24 same-gender conversations, but only 15 mixed-gender ones.\footnote{If all conversations in the corpus are significantly entraining on a particular feature, same-gender conversations and mixed-gender conversations would each get a maximum weighted percentage of 50.} 

Across all three feature sets, we generally find that there are equal or more same-gender conversations than mixed-gender conversations among those conversations that show evidence of entrainment on all lexical features except fillers (Table~\ref{lexical-gender}), turn-level convergence on acoustic-prosodic features (Table~\ref{ap-turn-conv-gender}), conversation-level proximity on all CSW features (Table~\ref{csw-conv-prox-gender}), and turn-level synchrony on amount of CSW (Table~\ref{csw-turn-synch-gender}). An example conversation snippet from a pair of interlocutors showing some of these entraining characteristics is the interaction in Figure~\ref{fig:1} between $S_{A1}$ and $S_{B1}$, who are both female, where we see lexical entrainment on a frequent word and proximity on presence and strategy of CSW. 

However, the opposite is true for conversations that show evidence of turn-level proximity on acoustic-prosodic features (Table~\ref{ap-turn-prox-gender}), turn-level proximity on binary presence and amount of CSW (Table~\ref{csw-turn-prox-gender}), and turn-level synchrony on presence and strategy of CSW. An example conversation snippet from a pair of interlocutors showing some of these entraining characteristics is the interaction in Figure~\ref{fig:1} between $S_{A2}$, who is female, and $S_{B2}$, who is male, where we see proximity on acoustic-prosodic features and presence and amount of CSW.

\begin{table}
\centering
\begin{tabular}{ |c|c|c|c|c| } 
 \hline
Feature & $\%_{w}$ (FF+MM) & $\%_{w}$ FM \\
\hline
Top-100 (corpus) & 50 & 50 \\
Top-25 (corpus) & 50 & 46.7 \\
Top-25 (conv) & 50 & 46.7 \\
Cues & 50 & 50\\
Fillers & 47.9 & 50 \\
Overall incl. OOVs & 50 & 46.7\\
Overall excl. OOVs & 41.7 & 30\\
 \hline
\end{tabular}
\caption{Weighted percentage ($\%_{w}$) of same-gender (FF+MM) and mixed-gender (FM) conversations among significantly entraining conversations, separated by lexical feature.}
\label{lexical-gender}
\end{table}

\begin{table} 
\centering
\begin{tabular}{ |c|c|c|c|c| } 
 \hline
Feature & $\%_{w}$ (FF+MM) & $\%_{w}$ FM \\
\hline
Min. pitch & 37.5 & 23.3 \\
Mean pitch & 41.7 & 23.3 \\
Max. pitch & 31.3 & 30 \\
SD pitch & 27.1 & 43.3 \\
Min. intens. & 29.2 & 36.7 \\
Mean intens. & 35.4 & 23.3 \\
Max. intens. & 33.3 & 33.3 \\
SD intens. & 27.1 & 43.3 \\
Jitter & 33.3 & 20 \\
Shimmer & 33.3 & 13.3 \\
HNR & 35.4 & 13.3 \\
SR & 31.3 & 16.7 \\
 \hline
\end{tabular}
\caption{Weighted percentage ($\%_{w}$) of same-gender (FF+MM) and mixed-gender (FM) conversations among significantly converging conversations at the turn-level, separated by acoustic-prosodic feature.}
\label{ap-turn-conv-gender}
\end{table}

\begin{table}
\centering
\begin{tabular}{ |c|c|c|c|c| } 
 \hline
Feature & $\%_{w}$ (FF+MM) & $\%_{w}$ FM \\
\hline
Min. pitch & 37.5 & 40 \\
Mean pitch & 39.6 & 50 \\
Min. intens. & 25 & 26.7 \\
Mean intens. & 35.4 & 36.7 \\
Max. intens. & 35.4 & 36.7 \\
HNR & 27.1 & 33.3 \\
SR & 35.4 & 33.3 \\
 \hline
\end{tabular}
\caption{Weighted percentage ($\%_{w}$) of same-gender (FF+MM) and mixed-gender (FM) conversations among significantly proximate conversations at the turn-level, separated by acoustic-prosodic feature.}
\label{ap-turn-prox-gender}
\end{table}

\begin{table}
\centering
\begin{tabular}{ |c|c|c|c|c| } 
 \hline
Feature & $\%_{w}$ (FF+MM) & $\%_{w}$ FM \\
\hline
CSW pres. & 45.8 & 33.3 \\
CSW amt. & 47.9 & 36.7 \\
CSW strat. & 39.6 & 33.3 \\
 \hline
\end{tabular}
\caption{Weighted percentage ($\%_{w}$) of same-gender (FF+MM) and mixed-gender (FM) conversations among significantly proximate conversations at the conversation-level, separated by CSW feature.}
\label{csw-conv-prox-gender}
\end{table}

\begin{table}
\centering
\begin{tabular}{ |c|c|c|c|c| } 
 \hline
Feature & $\%_{w}$ (FF+MM) & $\%_{w}$ FM \\
\hline
CSW pres. & 25 & 30 \\
CSW amt & 31.3 & 23.3 \\
CSW strat. & 27.1 & 30 \\
 \hline
\end{tabular}
\caption{Weighted percentage ($\%_{w}$) of same-gender (FF+MM) and mixed-gender (FM) conversations among significantly synchronous conversations at the turn-level, separated by CSW feature.}
\label{csw-turn-synch-gender}
\end{table}

\begin{table}
\centering
\begin{tabular}{ |c|c|c|c|c| } 
 \hline
Feature & $\%_{w}$ (FF+MM) & $\%_{w}$ FM \\
\hline
CSW pres. & 35.4 & 40 \\
CSW amt & 39.6 & 46.7 \\
 \hline
\end{tabular}
\caption{Weighted percentage ($\%_{w}$) of same-gender (FF+MM) and mixed-gender (FM) conversations among significantly proximate conversations at the turn-level, separated by CSW feature.}
\label{csw-turn-prox-gender}
\end{table}

Examining same-gender conversations more granularly, we notice that when there are more same-gender conversations than mixed-gender conversations among the significantly entraining conversations, either the proportion of FF conversations is greater than MM conversations or these are roughly equal within same-gender conversations. No such obvious pattern exists when there are more mixed-gender conversations than same-gender conversations among the significantly entraining conversations.\footnote{Within the set of mixed-gender conversations, female and male speakers were about equally likely to initiate CSW, with female interlocutors doing so in 8 of 15 such conversations.}

\section{Discussion}

From our experiments on the \textbf{lexical features} of code-switched conversations, we find statistically significant evidence of lexical entrainment across all of the word classes considered, as well as on overall language use within the corpus. 
We now briefly discuss why some speakers do not entrain to their interlocutors’ speech even though their interlocutors entrain to theirs. Prior research suggests that power differentials between speakers and other sociolinguistic factors tend to have an influence on such asymmetric entraining behavior, e.g. \citet{Danescu_Niculescu_Mizil_2011}. We suspect that L2 speaker proficiency could additionally play a role in this, although this has not yet been explored, to our knowledge, because existing data sets do not include this information. This is an interesting direction for further work, and we plan to collect the relevant demographic data for inclusion in our future studies of entrainment in code-switching between different language pairs. Overall, our findings are consistent with prior work on lexical entrainment in monolingual settings, e.g. \citet{weise-etal-2021-talk}, suggesting that the expected entraining behaviors previously observed in monolingual conversations also occur in code-switched settings, with potential implications for the cross-lingual nature of entrainment as a linguistic phenomenon.

Our experiments on \textbf{acoustic-prosodic features} provide evidence of entrainment in terms of turn-level proximity and convergence. As with the lexical features we investigated, this finding aligns with prior work on monolingual domains, e.g. \citet{levitan11_interspeech}, though we acknowledge that these authors additionally found varying degrees of significant entrainment in terms of conversation-level proximity and convergence and turn-level synchrony. We suspect that this difference might be due to the undirected nature of the data we use, unlike the task-oriented conversations \citeauthor{levitan11_interspeech} worked on. Nevertheless, we still see indications of patterns of \emph{local} entrainment in monolingual conversations generalizing to code-switched conversations.

We generally find consistent evidence of proximity at the turn- and conversation-level on the \textbf{CSW features} of the Bangor Miami conversations. On the presence of CSW, we find turn- and conversation-level proximity and turn-level synchrony. Though we find limited evidence of convergence at the turn- or conversation-level on CSW presence, our results suggest that speakers are entraining to one another in a way that surfaces as one speaker's choice to code-switch in speech encouraging the other to choose the same and vary in turn. These results are mirrored in our experiments on the amount of CSW, where we continue to observe that speakers seem to match one another's CSW behavior, specifically in terms of the quantity of code-switched utterances used within a turn, and the variation of the same across turns. Finally, from our experiments on CSW strategies, we find significant synchrony at the turn-level on insertional CSW, and significant proximity at the conversation-level on alternational CSW. The first of these findings mirrors the trend we see with entrainment on the presence of CSW. Since insertional CSW is a strategy lacking strict syntactic structure, when speakers mimic their interlocutors' variation in CSW in conversation, using insertional CSW is an easy way to achieve this kind of entrainment. This finding also aligns with that of \citet{ahn-etal-2020-code}, who similarly found this CSW strategy to show the most entrainment in human-machine written interaction. The second of these findings is more interesting, however, because alternational CSW requires knowledge of both languages' syntax and is by definition the most structured strategy for speakers to employ. This likely makes it easier for interlocutors to notice and mimic in comparison to insertional or other CSW. We particularly highlight this finding because significant evidence of entrainment on alternational CSW has not been found in previous work on written CSW, which suggests that there may be a key difference in how CSW and subsequent entrainment behaviors are produced in the spoken versus written modality. 

Finally, our preliminary \textbf{qualitative gender analysis} shows a slight skew toward entrainment taking place in same-gender conversations over mixed-gender ones. Compared to prior work that has considered the role of speaker gender in entrainment behavior, both \citet{BILOUS1988183} and \citet{levitan-etal-2012-acoustic} found more entrainment in mixed-gender pairs than same-gender pairs, but these were on task-oriented monolingual conversations. Our opposing results may 
suggest that the linguistic differences between task-oriented and undirected spontaneous speech (e.g. as detailed in \citet{pilán2023conversational}) carry through to differences in entrainment behaviors in these differing conversational settings, but further work is required to confirm or refute this. 

\section{Conclusion}

In this work, we have found that patterns of entrainment originally seen in monolingual domains are also present in a code-switched setting (\textbf{RQ1}). This is shown in particular by our experiments on lexical and acoustic-prosodic features of a corpus of code-switched conversations.
We also find evidence of entrainment on measures of CSW --- our work is the first to study entrainment in the context of spontaneous code-switched speech. While some patterns of entrainment previously found in code-switched written interactions are also present in spontaneous speech (\textbf{RQ2}), we have found additional ones that suggest a fundamental difference in CSW and corresponding entraining behaviors between modalities. We particularly find differences in the dimensions over which entrainment occurs in code-switched settings depending on the aspect of language production under consideration --- i.e. turn-level proximity and convergence on acoustic-prosodic features, versus turn- and conversation-level proximity on CSW features --- which points to the importance of examining multiple dimensions of entrainment when conducting analyses across feature sets. Our preliminary gender analysis similarly shows possible differences in entrainment between undirected and task-oriented speech. While we have uncovered these distinctions, we plan to further study and explain the reasons behind these in future work. 
Overall, these results provide promising implications for the potential "universal" nature of entrainment as a paralinguistic communication phenomenon that may be independent of monolingual versus multilingual language production.\footnote{We acknowledge that the findings of previous work such as \citet{levitan-developing} add nuance to the interpretation of our findings and to this claim, as entrainment is not guaranteed across all dimensions studied, even in monolingual data sets. However, we also point out that it is typical of entrainment studies to find evidence of entrainment along some, but not all, features under investigation (e.g. \citet{levitan11_interspeech}, \citet{weise-etal-2021-talk}, \citet{chen2023}), as in both \citet{levitan-developing} and our work.} We hope this work will serve as a stepping stone for informing the design of spoken language technology applications, particularly in the generation of more naturalistic and context-appropriate code-switched utterances by voice assistants. 

\section*{Limitations} 

In our work, we focused only on a single language pair, Spanish-English. Though we are excited about the potential implications of our findings, we acknowledge the need to extend our method to additional language pairs, ideally involving different language families, in order to test the robustness of our claim of generalizability. We intend to carry out this work in a future study. 

We also acknowledge that there are alternative ways of operationalizing certain measures of entrainment, e.g. synchrony as done in \citet{REICHEL201846}, as well as alternative frameworks of entrainment, e.g. the static versus dynamic entrainment framework of \citet{WYNN2022101173}. We chose to follow \citet{levitan11_interspeech} most closely for this first step towards understanding entrainment in CSW contexts, and will consider these alternatives in future work. 

Separately, it could have been interesting to include the semantic dimension of entrainment in our work, as did \citet{kejriwal23b_interspeech}. However, reliable embeddings are difficult to obtain for code-switched language. Future work might address this difficulty and investigate the incorporation of representations such as mBERT embeddings for this.  

\section*{Ethics Statement} 

This study was conducted on secondary data only, and did not require human experiments. We did not access any information that could uniquely identify individual users within the corpus, as its original creators had de-identified all speakers as outlined in the documentation of the data set. Though we did not collect the data used in this work, we note that all participants in the original corpus had explicitly consented to sharing the data that we analyze in our study. 

We believe this work is important because of its potential to inform the improvement of interactive speech technology that is capable of entraining to multilingual speakers in an appropriate and natural way, a quality of speech technology that users have been shown to prefer (e.g., as in \citet{BRANIGAN20102355}).

\newpage
\bibliography{anthology,custom}
\newpage
\appendix

\section{Appendix}
\label{sec:appendix}

\subsection{Lexical word classes}

The members of the affirmative cues word class are: \emph{alright, gotcha, huh, mm-hm, okay, right, uh-huh, yeah, yep, yes, yup, aja, claro, dale, ooh, sí, vale, venga}, as well as their spelling variants, e.g. \emph{uhuh} and \emph{aha} for \emph{uh-huh}, \emph{yah} for \emph{yeah}, etc. 

The members of the filled pauses word class are: \emph{ah, ahem, ay, eh, ehm, er, hmm, hmf, mm, pues, uff, uh, um}, as well as their spelling variants, e.g. \emph{mmm} for \emph{mm}, \emph{errr} for \emph{er}, etc. 

\begin{table} [h]
\begin{tabular}{ |c|c|c|c| } 
 \hline
Feature & \% of entraining convs (\emph{t}-, \emph{p}-val)\\ 
\hline
 \textbf{Min. pitch} & 
 \textbf{76.9 \quad(-2.14, 0.032)} \\
 \textbf{Mean pitch} & 
 \textbf{87.2 \quad(-3.69, 0.000)} \\ 
 Max. pitch & 
 84.6 \quad(-1.71, 0.088) \\ 
 SD pitch & 
 71.8 \quad(-0.928, 0.353)\\ 
 \textbf{Min. intensity} & 
 \textbf{51.3 \quad(-2.04, 0.041)} \\
 \textbf{Mean intensity} & 
 \textbf{71.8 \quad(-3.93, 8.49e-5)} \\
 \textbf{Max. intensity} & 
 \textbf{71.8 \quad(-4.45, 8.79e-6)} \\
 SD intensity & 
 69.2 \quad(-1.15, 0.249) \\
 Jitter & 
 89.7 \quad(-0.90, 0.37) \\
 Shimmer & 
 82.1 \quad(-1.46, 0.14)\\
 \textbf{HNR} & 
 \textbf{59.0 \quad (-2.61, 0.009)} \\
 \textbf{Speaking rate} & 
 \textbf{69.2 \quad(-2.25, 0.025)} \\
 \hline
\end{tabular}
\caption{Proximity at the turn-level. Significant acoustic-prosodic features are bolded.}
\label{tab:proximity-turn}
\end{table}

\end{document}